\documentclass{article}
\usepackage{spconf,amsmath,epsfig,graphicx,cleveref}

\usepackage{url}

\usepackage{todonotes}

\title{Biological Valuation Map of Flanders: A Sentinel-2 Imagery Analysis}
\name{Mingshi Li,$^1$ Dusan Grujicic,$^1$ Steven De Saeger,$^{3}$  Stien Heremans,$^{2,3}$ Ben Somers,$^{2}$ and Matthew B.\ Blaschko$^{1}$\thanks{We acknowledge funding from the Flemish Government
under FWO projects GEO.INFORMED S001421N and G0G2921N, and the Onderzoeksprogramma Artifici\"{e}le Intelligentie
(AI) Vlaanderen programme.}}
\address{ESAT-PSI$^1$ \& Dept.\ Earth and Environmental Sciences,$^2$ KU Leuven, Belgium\\
Instituut voor Natuur- en Bosonderzoek (INBO),$^3$ Belgium
}

\begin{document}
%
\maketitle
\begin{abstract}
In recent years, machine learning has become crucial in remote sensing analysis, particularly in the domain of Land-use/Land-cover (LULC). The synergy of machine learning and satellite imagery analysis has demonstrated significant productivity in this field, as evidenced by several studies \cite{dense_volpi_2017,deep_ma_2019,deep_yuan_2020,deep_zhu_2017}. A notable challenge within this area is the semantic segmentation mapping of land usage over extensive territories, where the accessibility of accurate land-use data and the reliability of ground truth land-use labels pose significant difficulties. For example, providing a detailed and accurate pixel-wise labeled dataset of the Flanders region, a first-level administrative division of Belgium, can be particularly insightful. Yet there is a notable lack of regulated, formalized datasets and workflows for such studies in many regions globally. This paper introduces a comprehensive approach to addressing these gaps. We present a densely labeled ground truth map of Flanders paired with Sentinel-2 satellite imagery. Our methodology includes a formalized dataset division and sampling method, utilizing the topographic map layout 'Kaartbladversnijdingen,' and a detailed semantic segmentation model training pipeline. Preliminary benchmarking results are also provided to demonstrate the efficacy of our approach.

\end{abstract}
\section{Introduction and background}
\label{sec:intro}

\subsection{Remote sensing with deep learning}
\noindent\textbf{Deep learning}
Remote sensing and satellite imagery analysis are key in geoscience and have expanded into urban planning, plantation coverage, and demographic monitoring. Mapping and evaluating land use globally presents a complex challenge, requiring interdisciplinary collaboration. Traditionally, this field relied on visual interpretation, spectral analysis, and the Normalized Difference Vegetation Index (NDVI). However, the advent of machine learning has revolutionized these methods. With the growing scale and complexity of satellite imagery data, machine learning significantly improves data processing capabilities, enhancing applications in various related fields. 

\noindent\textbf{Semantic segmentation, state-of-the-art models and challenges}
Semantic segmentation, a pivotal area in computer vision, involves assigning specific labels to each pixel in an image, a process that is crucial for a detailed understanding of images \cite{fully_shelhamer_2017}. This approach goes beyond object detection, which merely identifies object boundaries, and image classification, which categorizes an entire image under a single label. Semantic segmentation stands out by offering a detailed, pixel-level classification.

One of the most influential developments in this field is the introduction of the U-Net architecture by Ronneberger et al.\ (2015) \cite{unet_ronneberger_2015}. Initially used for biomedical image segmentation, U-Net has since become a cornerstone in various segmentation tasks due to its efficiency and effectiveness \cite{unet_zhou_2018,3d_iek_2016}. 
U-Net sets a benchmark in the wider field of image segmentation, which includes remote sensing research. 

While deep learning methods in remote sensing can provide promising results, it is widely acknowledged that developing reliable segmentation dataset with sufficient sample quantity for such training experiments poses a significant challenge due to its inherent complexity.

\subsection{Sentinel-2 database}
As part of the Copernicus Program, Sentinel-2A, operated by the European Space Agency (ESA), offers open-source earth observation imagery \cite{sentinel2_drusch_2012}. This satellite system provides high-resolution images in 13 spectral bands, covering visible, near-infrared, and shortwave infrared spectra with spatial resolutions of 10, 20, and 60 meters. 
Sentinel-2A's swath width of 290 km allows frequent coverage of Earth's land surfaces, crucial for monitoring environmental changes. For data processing, we utilized the OpenEO backend framework, which facilitates easy access and processing of Earth observation data \cite{rs13061125}. OpenEO provides a standardized interface to various Earth observation cloud backends, enhancing the accessibility of Sentinel-2A data products.

In our study, we used Sentinel-2A TOC (top of canopy) reflectance data from Terrascope, excluding the SCL (scene classification layer) for cloud masking. We focused on 11 data channels out of 13 for training our neural network model, initially based on the 10m resolution of visible light, with other channels resampled to 10m to serve as complementary features. This approach leverages the rich dataset of Sentinel-2A for advanced neural network training in environmental studies.

\section{Constructing the  BVM dataset}
\label{sec:format}

BVM (biological valuation map) is a densely labeled ground truth map describing the land use of the Belgian Flemish region \cite{618d3616ad2740ec910bf6677145828c,biological_saeger_2018}.\footnote{\url{https://www.vlaanderen.be/inbo/en-gb/biological-valuation-map/}} The original form of the dataset is different from what is ready for training use. Multiple steps were taken to construct such a semantic segmentation dataset.

\subsection{Polygon and label rasterization}
\textbf{Coordinate reference systems}
The BVM data was provided by the Flemish Research Institute for Nature and Forestry (INBO) in the format of a geographic shapefile containing labeled polygons covering the entire land area of Flanders. The coordinate reference system of these polygons is under the code EPSG:31370, the Sentinel-2 satellite imagery is provided in EPSG:32631. Projecting both systems to the most commonly used EPSG:4326 system solves the compatibility problem of the OpenEO backend.

\begin{figure*}[h]
  \centering
  \includegraphics[width=\textwidth]{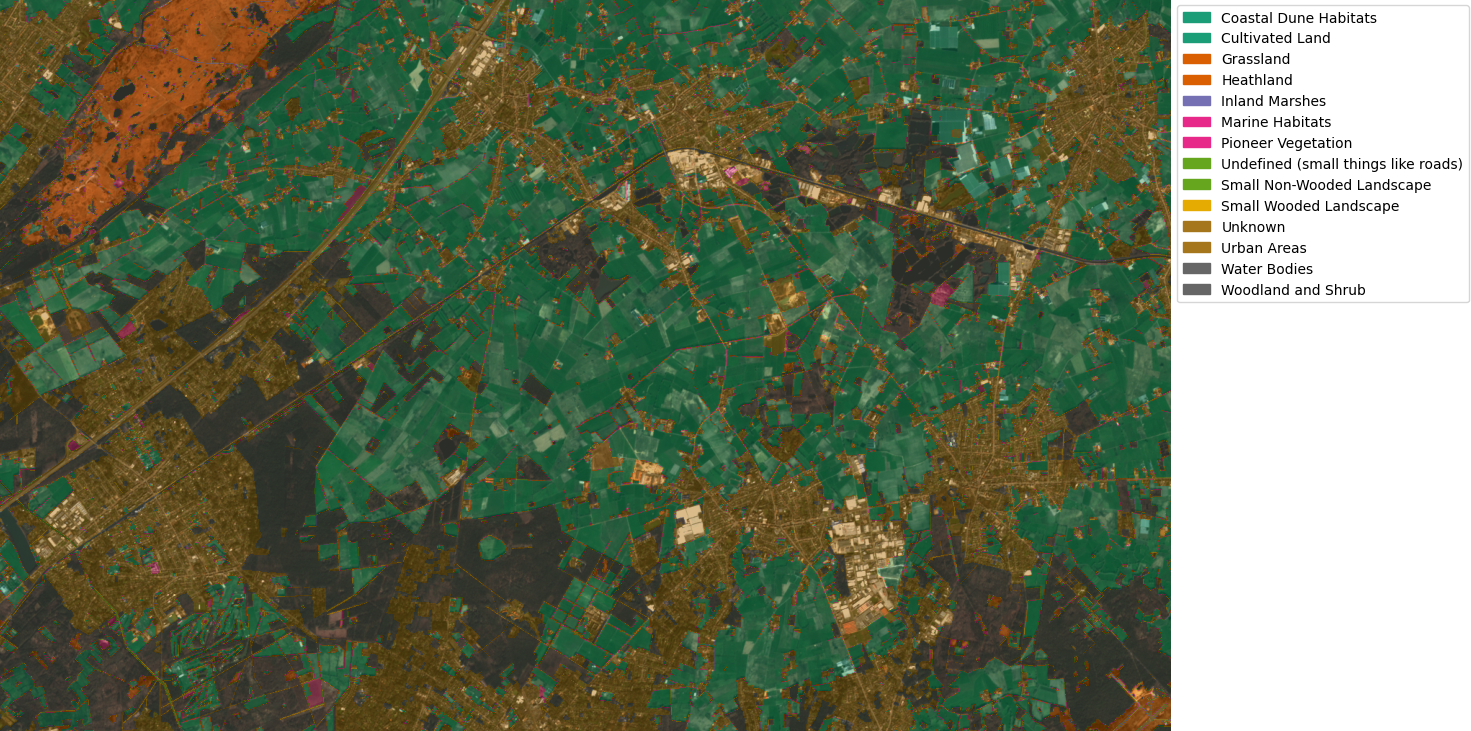}
  \caption{Example of a rasterized BVM map (different colors represent different pixel classes)}
  \label{fig:raster}
\end{figure*}

\noindent\textbf{Polygon rasterization}
In the BVM data map, all land areas are assigned labels in the form of polygons. To be used as ground truth data in our machine learning neural models, these polygons need to be rasterized so that each pixel on the BVM map is assigned to a class label individually as shown in \Cref{fig:raster}. There are in total 14 classes names listed in \Cref{tab:habitat_values_percentages}:

\begin{table}[h]
\small
\centering
\begin{tabular}{|p{5cm}|r|r|}
\hline
\textbf{Habitat Type} & \textbf{\# pixels} & \textbf{\%} \\ \hline
Coastal dune habitats & 183,904 & 0.14 \\ \hline
Cultivated land & 49,007,068 & 37.95 \\ \hline
Grasslands & 33,033,072 & 25.58 \\ \hline
Heathland & 1,112,397 & 0.86 \\ \hline
Inland marshes & 337,316 & 0.26 \\ \hline
Marine habitats & 329,994 & 0.26 \\ \hline
Pioneer vegetation & 940,283 & 0.73 \\ \hline
Small Landscape Features - not Specified & 81,046 & 0.06 \\ \hline
Small Non-woody Landscape Features & 192,329 & 0.15 \\ \hline
Small Woody Landscape Features & 925,387 & 0.72 \\ \hline
Unknown & 13,252 & 0.01 \\ \hline
Urban areas & 39,130,401 & 30.30 \\ \hline
Water bodies & 2,883,922 & 2.23 \\ \hline
Woodland and shrub & 16,657,189 & 12.90 \\ \hline
\end{tabular}
\caption{Habitat Types, Values, and Percentages}
\label{tab:habitat_values_percentages}
\end{table}

One can notice that the data label is unevenly distributed, with classes such as cultivated land and urban areas taking up the majority of the land area while classes like inland marshes and coastal dune habitats take up much less. We believe this class imbalance is often the case for most of the first-level administrative divisions in other countries too. Either inventing new training strategies or new metrics to address this problem in studies that adopt this type of dataset can be an interesting and challenging task.

\subsection{Pre-processing procedures}

In Flanders, satellite imagery acquisition often suffers from frequent cloudy weather, leading to heavily clouded Sentinel-2 data cubes. This significantly hinders model training, as cloudy pixels do not contribute useful features. To counter this, we utilized the SCL layer of Sentinel-2 data to mask out cloudy pixels, focusing solely on cloud-free pixels for training.

Additionally, the rectangular patches from Sentinel-2 sometimes extend beyond Flanders' borders. Since the BVM map only includes land-use data within Flanders, pixels outside this area are excluded, similar to cloudy pixels. To prevent data loss from patches partially outside Flanders, we implemented a strategy to mirror pixels near the Flemish border, creating a margin that ensures all pixels in a patch are usable for training. This approach maximizes data utility and maintains the integrity of our training dataset.

\subsection{Data preparation method}
\begin{figure*}[h]
  \centering
  \includegraphics[width=0.8\textwidth]{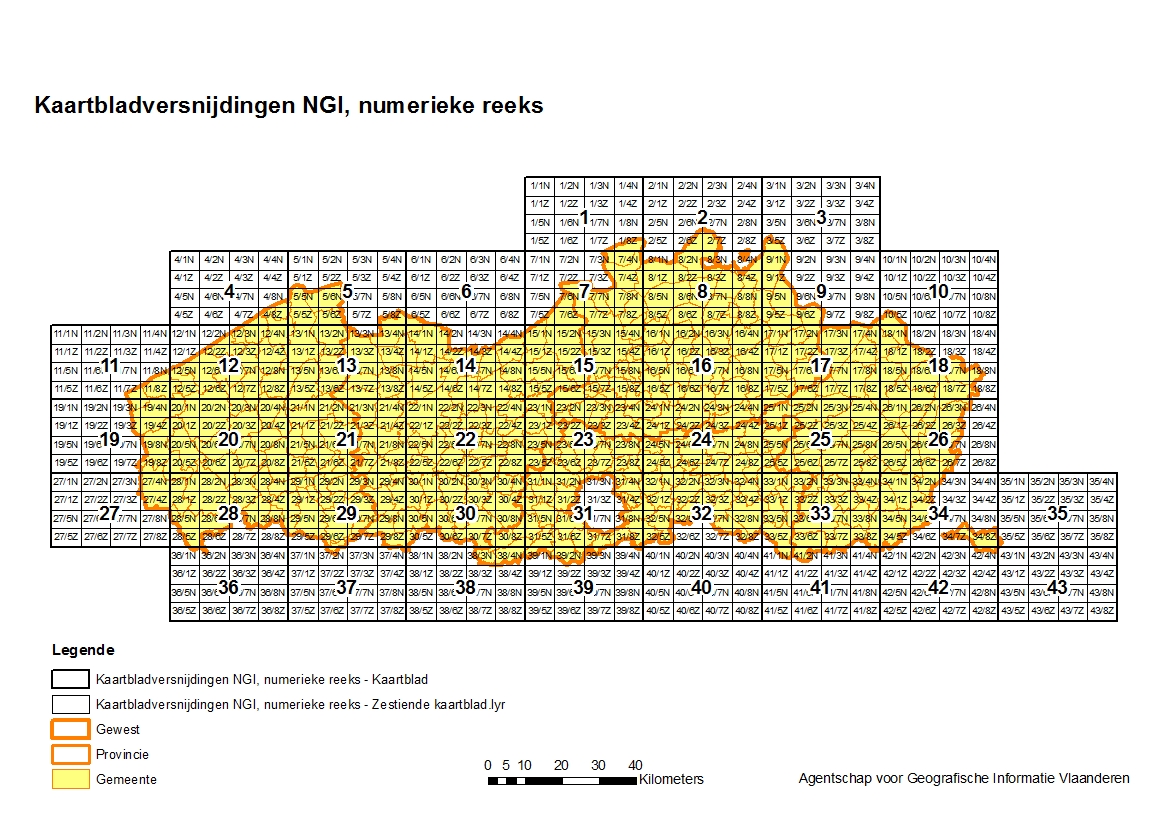}
  \caption{A demonstration of map sheet cutouts (Source: Digital Flanders Agency)}
  \label{fig:kaart}
\end{figure*}

In terms of preparation of the data, we introduce the topographic map sheet cutouts, the "kaartbladenversnijdingen (map sheet divisions)" (\Cref{fig:kaart}), to help us partition the BVM map further for training, validation, and test dataset splits. As the following figure shows: for each number X indexed block, we group X/1N, X/2N, X/1Z, X/2Z, X/5N, X/6N, X/5Z, X/6Z together as training blocks; X/13N, X/4N, X/3Z, X/4Z as validation blocks and X/17N, X/8N, X/7Z, X/8Z as test blocks. Each training or validation/test block covers a land area of size 160 $km^2$.

Using a kaartbladen grid to split up Flanders and choosing corresponding blocks is a robust way to partition training, validation, and test datasets. To ensure homogeneity in the class distribution of these three datasets, we report the Chi-squared distance between them. Chi-squared distance is often used to determine the similarity of two categorical distributions; it is formulated as follows:
$D_{\chi^2}(P, Q) = \sum_{i} \frac{(P(i) - Q(i))^2}{P(i) + Q(i)}$,
where P(i) and Q(i) are the probabilities of the i-th element in distributions P and Q, respectively.

The number of pixels for each class in three datasets and the total number of pixels are taken. The Chi-squared distance can thus be calculated as 0.0038 between training and validation sets, 0.00392 between training and test sets, and 0.0048 between validation and test sets. All three values are relatively low, indicating that the class distributions of all three sets are very similar to each other. Such close distances are desirable in our case as it suggests that the validation and test sets are good representatives of the training data. Details for class distribution are in \Cref{tab:class_distribution}. 

\begin{table}[h]
\centering
\small 
\begin{tabular}{|p{3cm}|r|r|r|}
\hline
\textbf{Class} & \textbf{Train (\%)} & \textbf{Val. (\%)} & \textbf{Test (\%)} \\ \hline
Coastal dune habitats & 0.10 & 0.06 & 0.25 \\ \hline
Cultivated land & 34.27 & 33.92 & 32.85 \\ \hline
Grasslands & 23.07 & 22.92 & 22.14 \\ \hline
Heathland & 0.54 & 1.08 & 0.95 \\ \hline
Inland marshes & 0.24 & 0.23 & 0.22 \\ \hline
Marine habitats & 0.26 & 0.04 & 0.34 \\ \hline
Pioneer vegetation & 0.69 & 0.65 & 0.56 \\ \hline
Small landscape features - not specified & 0.05 & 0.06 & 0.07 \\ \hline
Small non-woody landscape features & 0.14 & 0.13 & 0.12 \\ \hline
Small woody landscape features & 0.63 & 0.58 & 0.73 \\ \hline
Unknown & 0.01 & 0.007 & 0.006 \\ \hline
Urban areas & 26.27 & 27.05 & 28.56 \\ \hline
Water bodies & 2.08 & 2.05 & 1.74 \\ \hline
Woodland and shrub & 11.65 & 11.22 & 11.47 \\ \hline
\end{tabular}
\caption{Class distribution ratios in train, validation, and test Datasets}
\label{tab:class_distribution}
\end{table}

\section{Segmentation training tests}
\label{sec:pagestyle}

\subsection{Experiment specification}
\textbf{Model settings}
Our experiment is done on an Nvidia A100 40GB GPU with 256GB RAM, the model we adopted is vanilla U-Net with VGG16 as the encoder backbone, and the network is initiated with pre-trained weights on the Imagenet dataset \cite{imagenet_russakovsky_2015}. Depending on how many input channels are required in the input layer, the first layer of the network is modified correspondingly, where features from either 3 or 11 channels are merged together. The training is scheduled for 20,000 steps; the height and width of each training image are 256x256; and the batch size is set at 8.

\noindent\textbf{Dataloader}
Our dataloader is designed for handling Sentinel-2 satellite imagery data cubes in the Flanders region. It efficiently manages features like cloud coverage and region-specific data handling. The dataloader is initialized with user-defined parameters such as areas (kaartbladen), years, months, patch size, and data split (train, validation, test). Cloud masking using the SCL layer from Sentinel-2 data cubes has been applied. 
Region-specific processing is a key aspect, as the dataloader filters out data patches beyond Flanders' boundaries, maintaining alignment with the BVM map's geographical scope. It also employs a mirroring function for patches partially extending beyond Flanders' border.

Given the dataset size and convolution nature in the contracting path, where a target region is influenced by adjacent areas, our dataloader avoids simple checkerboard cutting of the BVM map to preserve relational information between nearby regions. It performs random cropping of satellite images and masks, ensuring spatial consistency in the cropped patches. This involves validating input types, applying padding, and determining cropping coordinates. This random cropping is vital for creating diverse training samples while maintaining the map's overall class distribution. The details of our project, including the dataloader and instructions on how to use it, can be found at \url{https://github.com/limingshi1994/GEOInformed/}.

\subsection{Results discussion}

Our experiments were conducted in three modes: 3-channel RGB, 3-channel RGNIR, and 11-channel, using different band combinations. The 3-channel RGB mode utilizes bands B02, B03, B04, while 3-channel RGNIR uses B03, B04, B08 (near infrared), and the 11-channel mode employs all available 11 bands. The network employs an early fusion mechanism where the initial convolution block merges features from all channels to produce feature embeddings in subsequent stages.

\begin{figure}[ht]
\centering
\includegraphics[width=\columnwidth]{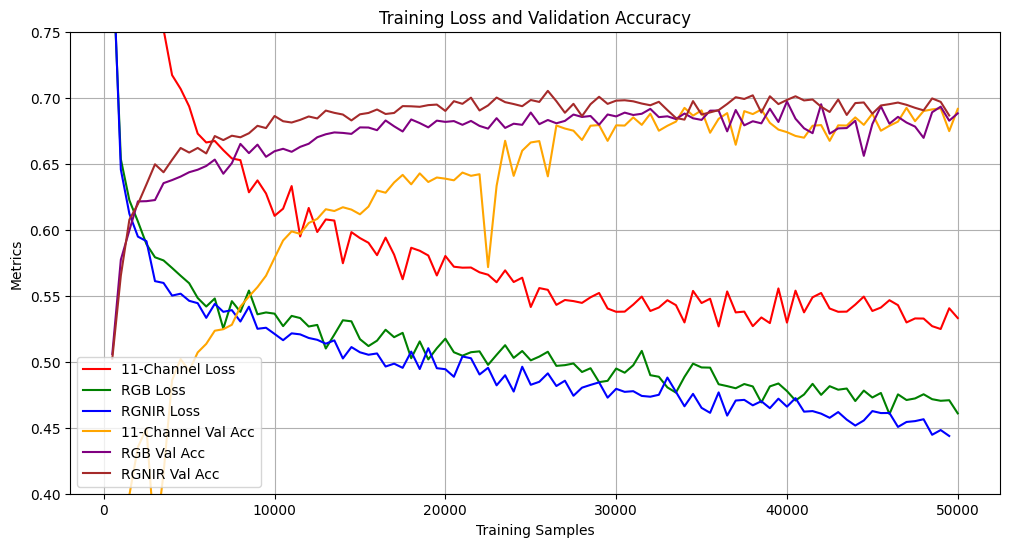}
\caption{Loss and accuracy of 3 modes}
\label{fig:losacc}
\end{figure}

In terms of metrics, we monitored training and validation losses and overall pixel accuracy (\Cref{fig:losacc}). Overall accuracy is determined by the ratio of correctly predicted pixels to all valid target pixels. Results indicate that the basic U-Net model, with a simple encoder architecture, achieves an average accuracy of 67\% for RGB input and 69\% for RGNIR input. The slight improvement with RGNIR input suggests that spectral bands beyond visible RGB light provide additional useful information for feature learning. Exploring more combinations, backed by remote sensing and geoscience theories, could potentially further enhance performance.

For the 11-channel input, we also achieved a 69\% accuracy, similar to RGB. This could be attributed to several factors. Firstly, our U-Net model is relatively simple, with only 23 million trainable parameters, which could be insufficient given our dataset's size and the complexity of temporal and spatial pixel features. Based off scaling laws proposed by Kaplan et al.\ for language models \cite{scaling_kaplan_2020}, we hypothesize that our model might have reached its performance limit simply inputting more features provided less improvement. Secondly, the feature fusion in our model occurs early via simple downsampling convolutions. While this reduces computational demands, it may not fully exploit the rich features of the 11-channel input. Researchers with more resources could develop more complex CNN architectures to better utilize the dataset's potential.

\section{Conclusion}
The results from our training experiments indicate that the BVM dataset, which we specifically developed for deep neural network training, meets the essential criteria for a comprehensive machine learning approach in the context of remote sensing studies. Firstly, we have contributed a densely labeled remote sensing semantic segmentation dataset, which covers an entire first-level administrative division of Belgium. The number of samples and credibility of labels makes it a valuable resource for future research. Secondly, we introduced a systematic method for dividing the dataset into training, validation, and test segments. This approach ensures a remarkable consistency across all three categories, which is crucial for the reliability and effectiveness of machine learning models. Thirdly, our initial experiments with this dataset, employing a basic, unoptimized U-Net model, yielded satisfying results, particularly in terms of overall accuracy. These findings underscore the dataset's potential even with relatively simple neural network models.

Additionally, the dataset presents several challenges that are pivotal in advancing model training towards enhanced performance. These challenges include addressing class imbalance, managing the substantial computational resources required, and the timing and methods of feature fusion in the contracting path of the network.

\bibliographystyle{IEEEbib}
\bibliography{untitled}

\end{document}